\documentclass[letterpaper, 10 pt, conference]{ieeeconf}
\IEEEoverridecommandlockouts
\usepackage{cite}
\usepackage{amsmath,amssymb,amsfonts}
\usepackage{algorithmic}
\usepackage{graphicx}
\usepackage{booktabs}
\usepackage{textcomp}
\usepackage{xcolor}
\usepackage[hyphens]{url}

\def\BibTeX{{\rm B\kern-.05em{\sc i\kern-.025em b}\kern-.08em
    T\kern-.1667em\lower.7ex\hbox{E}\kern-.125emX}}

\begin{document}

\title{Reducing Hallucinations in LLM-based Scientific Literature Analysis Using Peer Context Outlier Detection}

\author{
\authorblockN{Daniel Xie\authorrefmark{1},
Maxwell J. Jacobson\authorrefmark{1},
Adil Wazeer\authorrefmark{2},
Haiyan Wang\authorrefmark{2},
Xinghang Zhang\authorrefmark{2},
Yexiang Xue\authorrefmark{1}}
\authorblockA{\authorrefmark{1}Department of Computer Science, Purdue University, IN, USA\\
\authorrefmark{2}School of Materials Engineering, Purdue University, IN, USA\\
Email: \{xie476, jacobs57, awazeer, hwang00, xzhang98, yexiang\}@purdue.edu}
}

\maketitle
\thispagestyle{empty}
\pagestyle{empty}

\begin{abstract}
Reducing hallucinations in Large Language Models (LLMs) is essential for accurate data extraction from large text corpora. Current methods, like prompt engineering and chain-of-thought prompting, focus on individual documents and fail to consider relationships across a corpus. This paper introduces Peer Context Outlier Detection (P-COD), which uses inter-document relationships to improve extraction accuracy in scientific literature summarization, where papers with similar experiment settings should draw similar conclusions. By comparing extracted data to validated peer information within the corpus, we adjust confidence scores and flag low-confidence results for expert review. Our experiments demonstrate up to 98\% precision in outlier detection across 6 scientific domains, reducing hallucinations and letting researchers focus on genuinely ambiguous cases.
\end{abstract}


\begin{figure}[!tb]
    \centering
    \includegraphics[width=\linewidth]{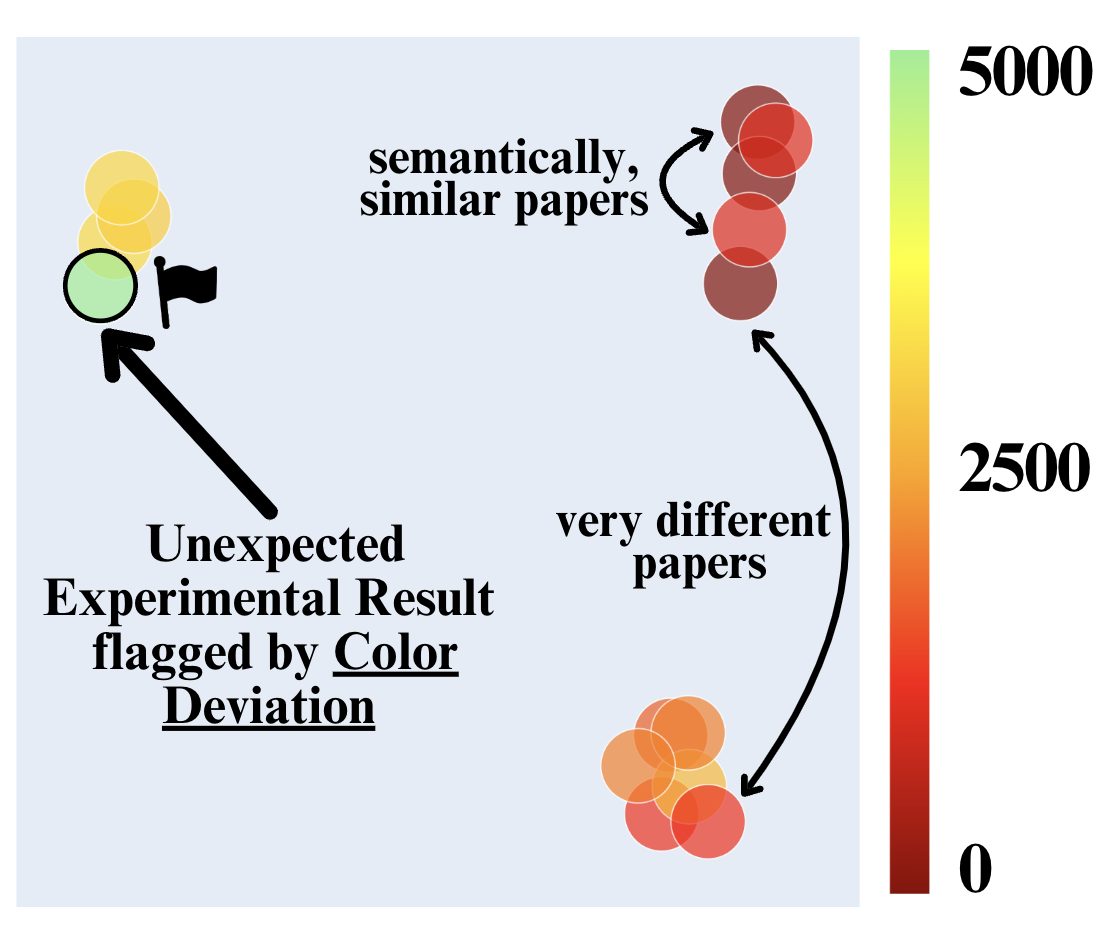}
    \caption{High-level idea behind Peer Context Outlier Detection. Each point is a scientific paper; color indicates an experimental result value. Semantically similar papers cluster together. The black flag marks an outlier whose extracted value deviates from its peers, suggesting a likely LLM hallucination to flag for human review.}
    \label{fig:peer-disagreement}
\end{figure}

\section{Introduction}

Large Language Models (LLMs) have transformed automated information retrieval by enabling rapid synthesis of scientific literature. However, their propensity for hallucinations -- plausible but incorrect information -- presents a serious challenge in scientific research, where accuracy is paramount. Hallucinations occur because LLMs generate responses from statistical patterns rather than verified facts, producing unreliable extractions that can mislead researchers and amplify misinformation in downstream analyses.

Several strategies aim to mitigate this. Prompt engineering \cite{Sahoo_2024,White_2023} guides the model toward more reliable extractions via zero- or few-shot prompts, but operates on individual documents and does not verify correctness against external sources. Chain-of-Thought (CoT) prompting \cite{wei2022chain,zheng2023ddcot} improves logical consistency through step-by-step reasoning but does not validate extractions against a corpus. Retrieval-Augmented Generation (RAG) \cite{He_2024,Lewis_2020,Gao_2023} grounds outputs in authoritative sources but depends on the completeness of its retrieval system.

Despite these advances, existing methods do not actively compare extracted data against related scientific literature; they fail to exploit \textbf{\textit{inter-document relationships}}. Without cross-document validation, plausible but incorrect outputs go undetected, researchers must manually verify every extraction, and erroneous findings can propagate into the literature.

To address this gap, we introduce \textbf{Peer Context Outlier Detection (P-COD)}. Unlike traditional methods that process documents in isolation, P-COD evaluates extracted data in the context of semantically related papers, under the intuition that studies with similar experimental conditions should yield similar results. P-COD provides confidence scoring based on alignment with peers, anomaly detection via numerical inconsistency across related papers, and human-in-the-loop validation so that only low-confidence results require manual review.

P-COD extends classical outlier detection \cite{Anusha_2021,Hodge_2004}, which typically relies on statistical measures such as Z-scores, Mahalanobis distance, or density-based methods like k-NN anomaly detection. These work well for structured numerical data but struggle on scientific extractions, where studies differ in methodology, reporting format, and experimental conditions -- a numeric outlier may actually be a valid deviation due to a legitimately different experimental approach. By grouping comparable studies through semantic embeddings and comparing extractions within those peer groups, P-COD reduces false positives and flags only values that are surprising \emph{relative to similar work}.

We evaluate P-COD on corpora spanning 6 domains (Computer Science, Physics, Biology, Chemistry, Materials Science, and Environmental Science), achieving up to \textbf{98\%} precision in identifying anomalous data points. The output is a ranked list scored by how surprising each data point is relative to peers, letting researchers prioritize review and dramatically reducing manual effort while improving reliability.

\section{Related Work}

Existing tools for LLM-based scientific data extraction \cite{10.1093/bioinformatics/btz682,gougherty2024testing,gao2021limitations} effectively parse text but lack robust validation. Semantic embedding models such as SciBERT measure textual similarity but do not assess the experimental consistency of extracted values.

Prompt engineering \cite{chen-etal-2022-adaprompt} uses carefully crafted prompts -- including zero-shot and few-shot instructions -- to guide more accurate extractions. Despite its effectiveness, it still treats each document in isolation, and unchecked errors propagate when results are taken at face value. Chain-of-Thought (CoT) prompting \cite{wei2022chain,zheng2023ddcot} improves logical consistency by forcing step-by-step reasoning, but does not validate extracted data against external sources or corpus-level trends \cite{turpin2023languagemodelsdontsay}.

Retrieval-Augmented Generation (RAG) \cite{Lewis_2020,Xiong_2024,Gao_2023,He_2024} supplements LLMs with an external retrieval system to cross-check extractions against authoritative sources. While this helps ground outputs, it depends on the completeness and coverage of the retrieval corpus, which may miss recent or niche scientific findings.

Recent surveys of large models catalog these failure modes more broadly: hallucination, limited factual grounding, and unreliable confidence are recurring limitations of large AI models across applications \cite{bi2025large}, and they compound at the scale of scientific literature analysis, where AI-related documents must be identified and mined in large corpora \cite{liang2025deepdiveai}. The problem is especially acute in agentic and multi-step pipelines, where an unflagged error in one document propagates through subsequent cross-document reasoning \cite{deng2025agentic}.

A central limitation across these approaches is that they do not flag or alert the user when a data point is uncertain or contradicts established knowledge; they often assume LLM outputs are definitive \cite{Li_2024} without assigning confidence. P-COD addresses this gap with a field-wide validation mechanism that compares extracted data across multiple documents, adjusting confidence based on peer consistency and surfacing surprising data points for human review.

\section{Problem Definition}

The challenge of accurate LLM-based extraction lies in the model's tendency to generate plausible but incorrect values. Automated pipelines lack robust mechanisms to verify extractions, creating risk in high-stakes domains like scientific research. A reliable system must not only extract efficiently but also identify which extractions require human verification -- the human-in-the-loop problem P-COD addresses.

The system takes a collection of scientific papers and a query about their contents. Each paper is processed to extract relevant information, and rather than treating extractions in isolation, the system evaluates them against other papers in the dataset so outputs are assessed by how well they align with broader field trends.

The output is a list of flagged papers and data points likely to contain hallucinated or inconsistent extractions. Each carries a \emph{surprising score} -- defined below from semantic similarity and experimental consistency -- so that values deviating from semantically related work receive higher scores. A tunable threshold controls how much deviation is tolerated before flagging, and the resulting ranking lets reviewers start with the most anomalous results. This transforms an unmanageable task -- validating all extracted data -- into a prioritized process that directs human effort only where uncertainty is highest.

\begin{figure*}[!tb]
    \centering
    \includegraphics[width=0.9\linewidth]{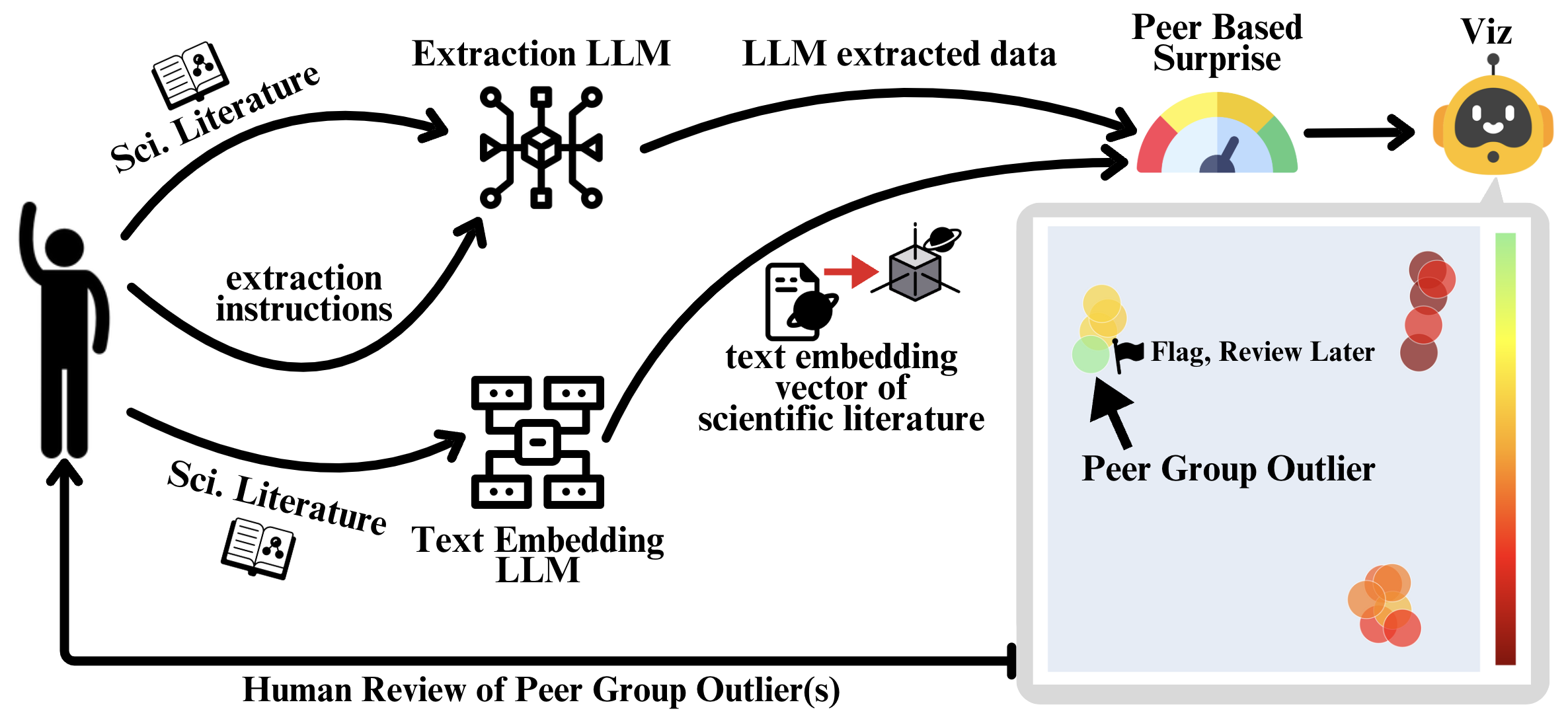}
    \caption{Workflow of P-COD. An extraction LLM pulls relevant data from papers; a text-embedding LLM places papers in a shared semantic space where proximity reflects methodological similarity. Within each peer group, extracted values are compared and a surprising score is assigned based on deviation. Papers whose results diverge sharply from close peers are flagged for human review; color deviation makes outliers immediately visible.}
    \label{fig:design_chart}
\end{figure*}

\section{Peer Context Outlier Detection (P-COD)}

P-COD identifies value outliers within a peer group by comparing extracted data points against semantically similar scientific papers. Traditional outlier detection struggles with contextual variation across studies, but our approach validates data relative to its peers (Fig.~\ref{fig:design_chart}), detecting whether a data point is expected \emph{within its research context} or anomalous. The method outputs a \textbf{\textit{Surprising Score}} quantifying how unexpected a data point is relative to its peers; high scores flag cases where content is semantically close to peers but experimental results diverge sharply.

\subsection{Semantic Text Embedding of Papers}

A text embedding is a high-dimensional vector representation that captures the semantic meaning of a text. Transformer-based LLMs produce these embeddings through multi-head self-attention, encoding both positional and contextual relationships between tokens. This allows comparison by meaning rather than lexical overlap -- unlike word-frequency baselines such as TF-IDF \cite{Ramos_2003}, which miss phrase composition and document-level semantics. We use OpenAI's \texttt{text-embedding-3-large} \cite{Sun_2024} to vectorize each paper. Papers with similar topics, methods, and terminology yield embeddings that are close in this space (Fig.~\ref{fig:design_chart}), enabling meaningful peer-group construction. We measure similarity using standard cosine similarity, the conventional choice for semantic vector comparison (defined in Eq.~\ref{eq:cosine_similarity}). This embedding-based approach enables context-aware evaluation: rather than examining an extracted value in isolation, we compare it against semantically related documents to determine whether it aligns with established findings in the field.

\subsection{Surprising Score Calculation}

The surprising score combines semantic similarity with experimental consistency. First, we compute the experimental difference score \(D_p\), the deviation of data point \(p\) across the selected numerical field:

\begin{equation}
D_p = \frac{|x_p - y_p|}{R_p}
\end{equation}

where \(x_p\) and \(y_p\) are the extracted and reference values for the same field, and \(R_p\) is the field's observed range (max minus min), normalizing deviations across measurement types.

The surprising score \(S_p\) sums over all neighboring points \(p' \in \mathcal{N}(p)\), \(p' \neq p\), in the semantic neighborhood of \(p\):

\begin{equation}
S_p = \sum_{p' \in \mathcal{N}(p)} w_v \cdot \mathrm{cos\_sim}(\text{emb}(p), \text{emb}(p')) + w_e \cdot D_p
\end{equation}

Here \(w_v\) and \(w_e\) weight semantic similarity and experimental deviation respectively, \(\text{emb}(\cdot)\) is the embedding, and \(D_p\) is the experimental deviation. The term \(\mathrm{cos\_sim}(\cdot,\cdot)\) is the standard cosine similarity between two embedding vectors,

\begin{equation}
\mathrm{cos\_sim}(u,v) = \frac{u \cdot v}{\lVert u \rVert \lVert v \rVert},
\label{eq:cosine_similarity}
\end{equation}

with no modification beyond this textbook definition. Higher \(S_p\) indicates greater deviation from expected results, surfacing anomalies that purely statistical or per-document methods miss.

\subsection{Threshold-Based Flagging}

Papers with $S_p > T$ are flagged for human review; $T$ is tunable to control the precision/recall trade-off.

\section{Experiments}

\begin{figure*}[!tb]
    \centering
    \includegraphics[width=0.95\linewidth]{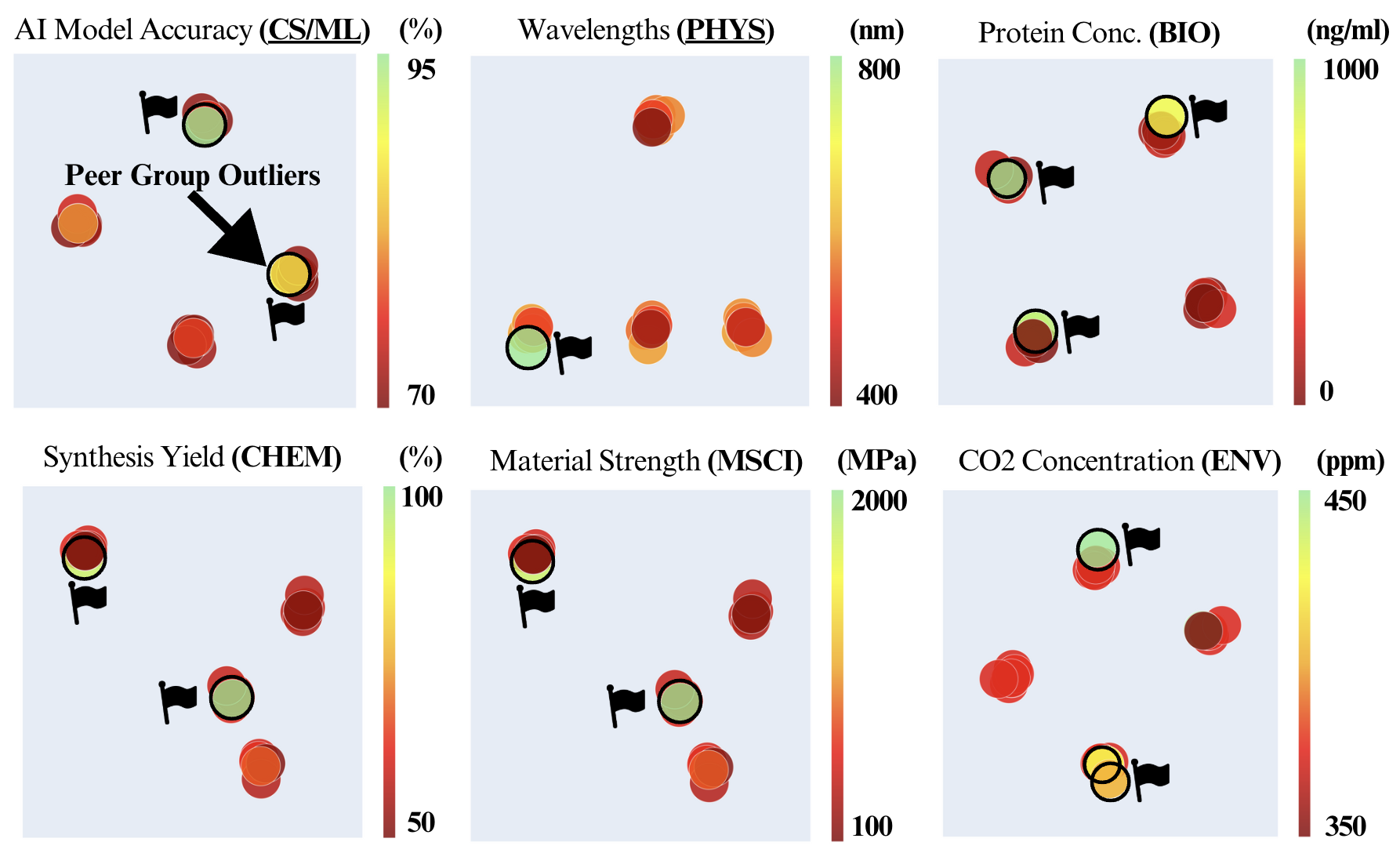}
    \caption{Multi-domain validation across six scientific fields. Each subplot shows semantic clustering and outlier detection with domain-specific metrics: CS (accuracy, 70--95\%), Physics (wavelength, 400--800\,nm), Biology (protein concentration, 2--1000\,ng/mL), Chemistry (yield, 50--100\%), Materials (tensile strength, 100--2000\,MPa), and Environmental Science (CO$_2$, 380--450\,ppm). Points are individual papers positioned by semantic similarity, colored by experimental value. Black flags mark synthetic corrupted papers that P-COD identifies as peer outliers.}
    \label{fig:multi_domain_results}
\end{figure*}

We conducted three experiments: (1) multi-domain validation across 6 fields, (2) large-scale single-domain testing, and (3) baseline comparisons. Results show consistent high performance, with up to 98\% precision in detecting corrupted numerical data while generalizing across domains.

\subsection{Experiment 1: Multi-Domain Validation}

\noindent\textbf{Setup.} To evaluate generalizability, we spanned 6 domains: Computer Science, Physics, Biology, Chemistry, Materials Science, and Environmental Science. Seed papers were drawn from the scientific\_papers dataset \cite{cohan2018discourse} (ArXiv and PubMed), which contains structured scientific documents. Using \texttt{text-embedding-3-large} to produce semantic vectors, we generated 168 synthetic papers (28 per domain) that maintain semantic similarity to their seeds while varying experimental values. These synthetic papers serve as ground truth, organized into 4 clusters per domain based on methodological similarity. Domain-appropriate metrics and typical ranges were identified through systematic analysis of the original dataset. Per-domain metrics were: accuracy (0.70--0.95) for CS, wavelength (400--800\,nm) for Physics, protein levels (2--1000\,ng/mL) for Biology, yield (50--100\%) for Chemistry, strength (100--2000\,MPa) for Materials, and CO$_2$ (380--450\,ppm) for Environmental. We strategically corrupted 25\% of papers by introducing values 3--5$\times$ outside normal ranges to create obvious outliers for validation.

\noindent\textbf{Results.} P-COD achieved consistent performance across all six domains. It flagged 53 papers; 42 were the truly corrupted ones, for 79.2\% precision (42/53) overall. The semantic clusters tightly grouped methodologically similar papers, and the surprising score consistently surfaced the corrupted values as outliers. This consistent performance across diverse scientific fields validates the domain-independent effectiveness of the approach.

\subsection{Experiment 2: Large-Scale Single-Domain Testing}

\noindent\textbf{Setup.} To test scalability and within-domain precision, we built a 200-paper CS dataset in 8 specialized clusters of 25 papers each (Deep Learning Optimization, Reinforcement Learning, Graph Neural Networks, Blockchain, Data Mining, Cloud Computing, Information Retrieval, and Bioinformatics). Each cluster had one seed and 24 synthetic papers sharing the accuracy metric (0.70--0.95); 25\% were corrupted with out-of-range accuracies.

\noindent\textbf{Results.} P-COD achieved 98.0\% precision, detecting 49 of 50 corrupted papers. This demonstrates that the method scales and maintains high precision even when semantic similarity is high across closely related sub-fields.

\begin{figure}[!htb]
    \centering
    \includegraphics[width=\columnwidth]{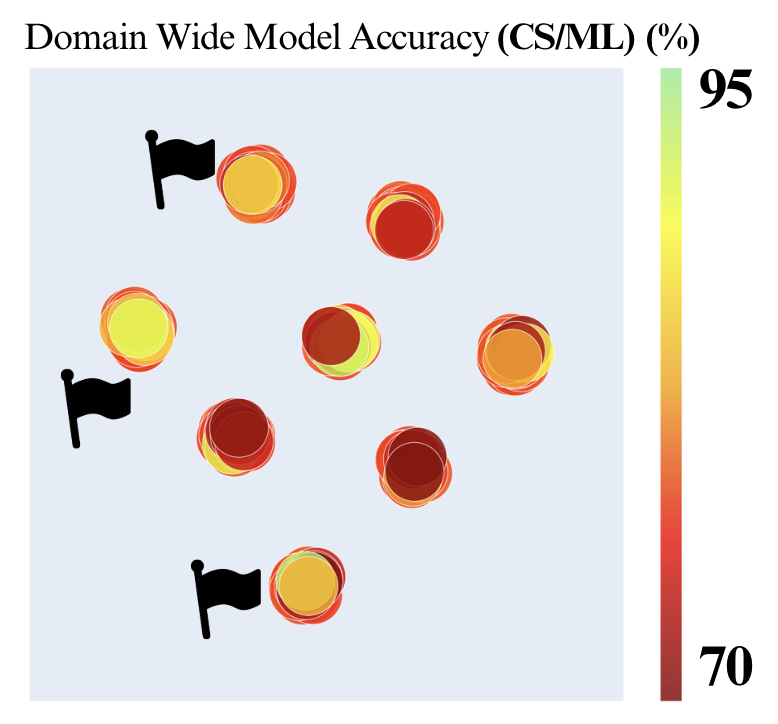}
    \caption{Large-scale single-domain clustering across 8 CS sub-fields (200 papers, shared accuracy metric 0.70--0.95). Cluster separation is increased (factor 3.0) for visualization; black flags mark a representative sample of papers P-COD identified as outliers. Tight clustering shows CS papers are semantically closer than cross-domain papers, yet P-COD still precisely detects experimental anomalies within them.}
    \label{fig:cs_clustering}
\end{figure}

\begin{table*}[!tb]
\centering
\renewcommand{\arraystretch}{1.2}
\setlength{\tabcolsep}{10pt}
\caption{Precision of outlier detection methods across six scientific domains. CS is from the large-scale single-domain experiment.}
\label{tab:multi_domain_supercolumn}
\begin{tabular}{lcccccc}
\toprule
\multicolumn{1}{c}{} & \multicolumn{6}{c}{\textbf{Precision (\%)}} \\
\textbf{Method} & \textbf{CS} & \textbf{Physics} & \textbf{Biology} & \textbf{Chemistry} & \textbf{Materials} & \textbf{Env. Sci.} \\
\midrule
P-COD (Ours) & \textbf{98.0} & \textbf{66.7} & \textbf{80.0} & \textbf{100.0} & \textbf{87.5} & \textbf{100.0} \\
Prompt Engineering & 52.3 & 48.1 & 63.8 & 57.2 & 45.7 & 68.4 \\
Chain-of-Thought & 61.7 & 58.3 & 74.2 & 67.9 & 62.4 & 76.8 \\
\bottomrule
\end{tabular}
\end{table*}

\subsection{Experiment 3: Baseline Comparisons}

\noindent\textbf{Setup.} We compared against two standard LLM-based approaches. The \textbf{prompt engineering baseline} uses direct GPT-4 classification with domain-specific prompts combining range checking and title heuristics. The \textbf{Chain-of-Thought baseline} performs structured multi-step reasoning (range assessment, deviation severity, contextual analysis, and domain validation) with weighted scoring for the final decision. Both include fallback mechanisms for consistency.

\noindent\textbf{Results.} P-COD significantly outperformed both baselines across every domain (Table~\ref{tab:multi_domain_supercolumn}). Prompt engineering achieved 45.7--68.4\% precision (avg.\ 55.9\%); CoT achieved 58.3--76.8\% (avg.\ 66.9\%); P-COD achieved 66.7--100\% (avg.\ 88.7\%). The 98.0\% CS figure reflects the large-scale single-domain experiment.

The gap reveals that methods relying on per-document reasoning or statistical thresholds cannot capture the cross-paper relationships P-COD leverages through semantic peer grouping. Prompt engineering and CoT guide the model's reasoning but cannot cross-reference related literature, so domain-valid deviations and genuine hallucinations are treated identically. P-COD's semantic similarity term is the distinction: a value that departs from close peers is far more suspicious than one that departs from a mean computed across methodologically diverse studies. The ``color-pop'' visualization further adds immediate interpretability that statistical methods lack.

\subsection{Discussion}

Across all three experiments, P-COD's strongest precision appears where peer groups are dense and methodologically homogeneous -- 98\% on the 200-paper CS dataset, 100\% on Chemistry and Environmental Science. The key practical takeaway is that the surprising score is a ranking tool: even when absolute precision varies, the highest-scoring extractions are reliably worth human review first, which is the operational guarantee the method was designed to provide. Computationally, P-COD scales favorably: embedding each paper is a one-time cost linear in corpus size, and the dominant runtime expense is the pairwise comparison within each semantic neighborhood rather than across the entire corpus, so cost stays manageable as the corpus grows. The 200-paper single-domain run (Experiment 2) confirms that precision holds at this larger scale.

P-COD rests on the assumption that a paper's semantic neighbors are valid experimental peers -- that embedding proximity is a reliable proxy for methodological comparability. This holds best in homogeneous fields and weakens where a single domain spans diverse methodologies: semantically similar papers may report legitimately different values because they use different experimental approaches, not because either is hallucinated. In such cases the experimental-deviation term \(D_p\) cannot yet separate a genuine error from a valid methodological divergence, so both inflate the surprising score and precision falls -- visible in the smaller, more heterogeneous domains (e.g. 66.7\% on Physics). P-COD therefore underperforms when peer groups are small, sparse, or methodologically heterogeneous, and is best deployed as a prioritization layer paired with human review rather than as an autonomous filter. Separating valid divergence from error -- for instance by conditioning \(D_p\) on experimental metadata -- is a direction for future work.

\section{Conclusion}

We presented Peer Context Outlier Detection, a field-wide validation method that combines semantic similarity with experimental consistency to flag hallucinated or inconsistent LLM extractions. By enabling cross-document validation, P-COD ensures that extracted data aligns with established trends within a given research domain, reducing the likelihood of hallucinations and improving the reliability of information synthesized from LLMs.

P-COD achieves up to 98\% precision in detecting anomalous data points across six scientific domains, enabling researchers to focus review effort where it matters most while minimizing unnecessary manual verification. This makes it practical to deploy in the scientific community to improve the reliability of automated literature analysis.

Future work will explore domain-specific adaptation through specialized embedding models for individual scientific fields, dynamic threshold optimization that learns flagging criteria from data, and multi-modal integration to analyze figures and tables alongside textual content. Temporal analysis could additionally help distinguish genuine scientific advances from extraction errors. These extensions build on our validated approach to further enhance accuracy and applicability.

\section*{Acknowledgment}
This research was supported by NSF Career Award IIS-2339844, DOE – Fusion Energy Science grant: DE-SC0024583.

\bibliographystyle{IEEEtran}
\bibliography{pcod_refs}

\end{document}